\documentclass[a4paper,fleqn]{cas-dc}

\usepackage[authoryear]{natbib}
\usepackage{graphicx}%
\usepackage{multirow}%
\usepackage{amsmath,amssymb,amsfonts}%
\usepackage{amsthm}%
\usepackage{mathrsfs}%
\usepackage[title]{appendix}%
\usepackage{xcolor}%
\usepackage{textcomp}%
\usepackage{manyfoot}%
\usepackage{booktabs}%
\usepackage{algorithm}%
\usepackage{algorithmicx}%
\usepackage{algpseudocode}%
\usepackage{listings}%
\usepackage{makecell}
\usepackage{tabularx}

\newcolumntype{B}{>{\hsize=.6\hsize}X}
\newcolumntype{M}{>{\hsize=.36\hsize}X}
\newcolumntype{s}{>{\hsize=.25\hsize}X}
\newcolumntype{V}{>{\hsize=.32\hsize}X}
\newcolumntype{Y}{>{\hsize=.45\hsize}X}
\newcolumntype{T}{>{\hsize=.2\hsize}X}

\begin{document}
\let\WriteBookmarks\relax
\def\floatpagepagefraction{1}
\def\textpagefraction{.001}
\shorttitle{Generative AI in Higher Education}
\shortauthors{Hui Wang et~al.}

\title [mode = title]{Generative AI in Higher Education: Seeing ChatGPT Through Universities' Policies, Resources, and Guidelines}                      

\author[1]{Hui Wang}[orcid=0009-0005-0226-3554]
\cormark[1]
\ead{hwang0524@arizona.edu}


\affiliation[1]{organization={Second Language Acquisition and Teaching, University of Arizona},
                city={Tucson},
                state={AZ},
                postcode={85721},
                country={USA}}

\author[1]{Anh Dang}

\author[2]{Zihao Wu}[orcid=0000-0001-7483-6570]
\cormark[1]
\ead{zihao.wu1@uga.edu}


\affiliation[2]{organization={School of Computing, University of Georgia},
                city={Athens},
                state={GA},
                postcode={30602},
                country={USA}}

\author[3]{Son Mac}

\affiliation[3]{organization={Department of Electrical \& Computer Engineering, University of
Arizona},
                city={Tucson},
                state={AZ},
                postcode={85721},
                country={USA}}

\cortext[cor1]{Corresponding authors}



\begin{abstract}
The advancements in Generative Artificial Intelligence (GenAI) provide opportunities to enrich educational experiences, but also raise concerns about academic integrity. Many educators have expressed anxiety and hesitation in integrating GenAI in their teaching practices, and are in needs of recommendations and guidance from their institutions that can support them to incorporate GenAI in their classrooms effectively. In order to respond to higher educators’ needs, this study aims to explore how universities and educators respond and adapt to the development of GenAI in their academic contexts by analyzing academic policies and guidelines established by top-ranked U.S. universities regarding the use of GenAI, especially ChatGPT. Data sources include academic policies, statements, guidelines, and relevant resources provided by the top 100 universities in the U.S. Results show that the majority of these universities adopt an open but cautious approach towards GenAI. Primary concerns lie in ethical usage, accuracy, and data privacy. Most universities actively respond and provide diverse types of resources, such as syllabus templates, workshops, shared articles, and one-on-one consultations focusing on a range of topics: general technical introduction, ethical concerns, pedagogical applications, preventive strategies, data privacy, limitations, and detective tools. The findings provide four practical pedagogical implications for educators in teaching practices: accept its presence, align its use with learning objectives, evolve curriculum to prevent misuse, and adopt multifaceted evaluation strategies rather than relying on AI detectors. Two recommendations are suggested for educators in policy making: establish discipline-specific policies and guidelines, and manage sensitive information carefully.
\end{abstract}




\maketitle

\section{Introduction}\label{sec1}

The development of artificial intelligence (AI) led to the rapid advancement of large language models such as ChatGPT, GPT-4, Gemini, Claude 2, Llama 2, and etc \citep{achiam2023gpt,team2023gemini,touvron2023llama}. In November 2022, OpenAI first released ChatGPT\footnote[1]{https://openai.com/blog/chatgpt}, which is a powerful language model-based chatbot that can understand human conversation and generate human-like texts. Since its release, ChatGPT has gained significant attention as well as vigorous discussion across a wide range of fields. In educational contexts, it can be employed through various applications, including generating ideas, revising grammatical errors, providing instant feedback, and evaluating and grading writing assignments \citep{abdullayeva2023impact,fuchs2023exploring,rudolph2023chatgpt}. However, the automatic generation of human-like texts also poses potential risks to academic integrity, especially when faced with writing-intensive assignments and language courses \citep{perkins2023academic,sullivan2023chatgpt}. Some scholars express additional concerns about potential misuse by students, suggesting that students may rely heavily on ChatGPT, which might further impact their critical thinking and problem-solving abilities \citep{kasneci2023chatgpt}. Due to the nature of a new and emerging technology with constant changes and updates, many educators have also expressed anxiety and hesitation in integrating GenAI in their teaching practices, and are in need of recommendations and guidance from their universities that can support them to implement GenAI in classrooms effectively. Thus, the question now forms from not only understanding what ChatGPT can do but also from what universities can offer and what faculty can apply in terms of guidance and strategies on the use of ChatGPT in educational academia. Specifically, it is necessary to examine how different universities and educators are currently perceiving, adapting to, and applying the use of such technology in higher education.

This study aims to investigate policies, guidelines, and resources currently provided by U.S. universities for their educators, teacher trainers, students, and researchers to adopt GenAI, especially ChatGPT, in their teaching, learning, and research in higher education. The findings will address numerous educators’ needs by informing AI-assisted teaching practices and guiding future policy-making, and thus impact the application of GenAI in higher education.

\section{Literature review}\label{sec2}

Generative Artificial intelligence (GenAI) and natural language processing (NLP) have emerged as groundbreaking technologies that rapidly attract worldwide attention in various fields \citep{kalla2023study,ray2023chatgpt,rice2024advantages}. ChatGPT, a revolutionary technology created by OpenAI, is an advanced chatbot that uses AI and NLP techniques to generate coherent and human-like responses \citep{kalla2023study}. Using deep learning and neural networks, this technology is equipped to understand, analyze, and produce responses to a wide variety of prompts, including questions, statements, or academic inquiries, all within a few seconds. 

GenAI has been increasingly affecting higher education, as it has the potential to enhance learning experiences while also posing challenges to the current educational contexts \citep{dempere2023impact,grassini2023shaping,onal2023cross}. It encourages students to ask questions, clarify their needs, and delve into various topics as a self-regulated learning approach \citep{chiu2023impact,cooper2023examining,rasul2023role,wu2024promoting}. In the study conducted by \citet{ng2024empowering}, students were taught to use ChatGPT at home to learn science concepts. The chatbot served as a resource to provide recommendations with science-related examples and explanations as well as scaffold the learning process by setting goals, suggesting learning strategies, and promoting time management skills \citep{ng2024empowering}. Another frequent pedagogical application of GenAI is as a writing assistant in teaching and learning academic writing \citep{crompton2023artificial,dempere2023impact,imran2023analyzing}. They can assist student writers during the planning, drafting, and revising phases of academic writing, and provide suggestions that address their writing needs from linguistic nuances to genre-specific features \citep{alharbi2023ai,liu2024investigating,mahapatra2024impact,yan2023impact}. As GenAI tools have been widely available for academics and students, concerns regarding academic integrity and copyright infringement become prominent in the discussions \citep{eke2023chatgpt,gao2023comparing,baek2023chatgpt,peres2023chatgpt}. Adopting AI-generated content without critical evaluation, rephrasing, and citation constitutes plagiarism and cannot be acceptable \citep{eke2023chatgpt,jarrah2023using}. In addition, some students experienced a lack of critical thinking and a decline in learning motivation due to overreliance on these tools \citep{escalante2023ai,harunasari2022examining,mahapatra2024impact,song2023enhancing}. Therefore, the ongoing conversation highlights the misuse issues as a great concern that should be addressed. Institutions as well as teachers of higher education need to examine, adopt, and revise policies and strategies in their teaching contexts to preserve academic integrity, prevent plagiarism, and promote ethical implementation of GenAI tools.  

While there is an increasing number of research experimenting with new technologies in education, there will always be anxiety and hesitation coming from teachers, especially due to the complexity and ethical issues of GenAI \citep{barrett2023not}. Studies by \citet{iqbal2022exploring} and \citet{kiryakova2023chatgpt} reported that a large portion of educators considered ChatGPT both a threat and a favorable opportunity and were cautious in using them. From the interviews in the study conducted by \citet{iqbal2022exploring}, many teachers pointed out that there was a lack of training and support on the effective use of GenAI from their institutions, which was identified as a major obstacle to its application. Similarly, \citet{barrett2023not} emphasized that deficiencies in GenAI policies and guidances from their institutions have led some instructors to hesitate to fully embrace GenAI due to ethical concerns and a naivety of how to appropriately use GenAI for educational purposes. Participating faculty in \citet{van2024chatgpt} advocated that more resources, such as webinars, should be provided for faculty and students to raise GenAI awareness and literacy. Overall, teachers’ perspectives highlight the importance and urgency of examining available policies and resources on GenAI use in higher education, which will guide institutions in providing effective guidelines and preparing instructors for future curriculum design. 

Standing in the midst of heated debates about the ethical use of GenAI in education, numerous studies are calling for further research on principles, strategies, and resources to harness GenAI as a value-driven opportunity to enhance learning \citep{ali2023let,firat2023chatgpt,lee2023multimodality,michel2023challenges,van2024chatgpt}. However, limited research have focused on exploring how higher education institutions are currently perceiving, adapting to, and applying the use of GenAI. This study fills the research gap by exploring publicly available policies, statements, guidelines, and resources regarding GenAI provided by the top 100 universities across the U.S., and from there, discussing implications and suggestions for educators in policy-making and teaching practices. 

\section{Research questions}\label{sec3}

\begin{itemize}
    \item How do U.S. universities manage and regulate the use of ChatGPT and other GenAI tools in higher education contexts, and what perceptions can be found from their policies on integration?
    \item What resources and guidance do U.S. universities provide on the use of ChatGPT and other GenAI tools for teaching practices?
    \item How do the school ranking tiers and academic specializations influence the trends in perceptions and resource provision regarding the use of GenAI in higher education and what pedagogical implications can we learn from them? 
\end{itemize}

\begin{figure}[ht]
\centering
\includegraphics[width=\columnwidth]{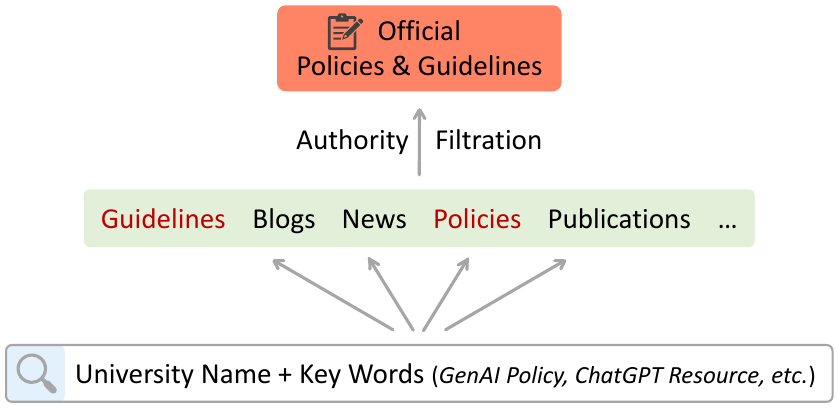}
\caption{Data collection and filtration process.}\label{data}
\end{figure}

\begin{table*}[ht]
\caption{Coding scheme for analyzing university policies and statements.}\label{tab1}
\begin{tabularx}{\textwidth}{@{}sMB@{}}
\toprule
Parent Codes                            & Child Codes                & Definition\\ \midrule
\multirow{7}{*}{University Decision} & Undecided/Unclear          & The university has not made a clear decision or taken a definitive stance regarding GenAI.\\ \cmidrule(l){2-3}
                                        & Allow use with Conditions  & The university permits the use of GenAI with conditions, such as appropriate citations.\\ \cmidrule(l){2-3}
                                        & Ban Use                    & The university prohibits the use of GenAI.\\ \cmidrule(l){2-3}
                                        & Instructor Decides         & The university allows the use of GenAI depending on the instructor’s decisions.\\ \midrule
\multirow{6}{*}{Instructor Decision}     & Prohibition by Default     & The use of GenAI is generally not allowed unless explicitly permitted by the instructor.\\ \cmidrule(l){2-3}
                                        & Permissibility by Default  & The use of GenAI is generally allowed unless explicitly prohibited by the instructor.\\ \cmidrule(l){2-3}
                                        & Neutral                    & The university relies on the instructor’s decision without a specific stance.\\ \midrule
\multirow{7}{*}{Education Purpose}    & Plagiarism Prevention      & To prevent students from directly copying texts generated from GenAI.\\ \cmidrule(l){2-3}
                                        & Authorship and Attribution & To require acknowledge AI-generated content in student academic assignments.\\ \cmidrule(l){2-3}
                                        & Limitations                & To address limitations, including biased, inaccurate, unreliable, or falsely cited information generated by AI. \\ \midrule
\multirow{4}{*}{Research Purpose}     & Intellectual Property      & To highlight the importance of acknowledging  AI-generated content in professional research settings.\\ \cmidrule(l){2-3}
                                        & Data Privacy and Security  & To address the confidentiality and security of data when using GenAI in professional research.\\ \bottomrule
\end{tabularx}
\end{table*}

\section{Methods}\label{sec4}
\subsection{Data Collection}
The data in this study consists of policies, statements, resources, and guidelines regarding the use of GenAI, especially ChatGPT, from the top 100 U.S. universities listed in the 2024 US News Best National University Rankings. These universities represent a broad spectrum of reputational educational institutions across the U.S. To collect the data, we first identified the top 100 universities (n = 104) in the U.S. via the 2024 US News Best National University Rankings. Then we performed a systematic search using a list of keywords (e.g., GenAI policy, GenAI guidelines) together with the name of each university. The search results were filtered by the sources and content for each text. They were evaluated by the research team according to the inclusion and exclusion criteria (see Fig. \ref{data}). The inclusion criteria contain: 1) data from official university sources, such as the Office of Provost, Academic Senate, Center of Teaching and Learning, and Library Resources; 2) university-wide policies and statements regarding the use of GenAI tools and academic integrity; 3) guidelines and resources in relation to the use of ChatGPT and other GenAI tools in teaching, learning, and research. The exclusion criteria include: 1) articles from online news, and blog posts; 2) sources from a specific department or program of each university. 

This study specifically focuses on data from official university sources. Firstly, official sources can provide authoritative insights that can be representative of the university's institutional stance and strategic direction. These policies and guidelines often come from the discussion by each university’s policymakers and/or official committee of faculty, staff, and students, who are the authority and/or representatives of the universities. Additionally, these official sources are provided directly by the universities and accessible to both internal and external members of the school community, which may indicate the university's endorsement and support for the public information on their official websites. Furthermore, a number of universities often regularly update the information presented on their websites based on their individual contexts and people’s opinions. Therefore, we believe that the policies from the official university sources can reflect a current and clear picture of how these universities perceive the new technologies, and the guidelines from the official sources can offer some valuable insights into how the universities were guiding and supporting their faculty and students to appropriately integrate GenAI.

It is important to note that the policies and guidelines collected in this study are up until April 2024. We acknowledge that the development of GenAI policies and resources is dynamic and there will be changes regarding these policies in the future, however, as for our study, the current dataset serves as the foundation for our analysis. For those interested in exploring this dataset further, it is accessible at \href{https://docs.google.com/spreadsheets/d/1OYlAN-R5t0dagEtBjQyfYztQkfGyahkG5sNNPx7HTTU/edit?usp=sharing} {the dataset of university policies and resources}.

\subsection{Coding Schemes} 
Thematic analysis \citep{braun2006using} was conducted to identify the themes related to universities’ perceptions and the availability of provided resources. Table \ref{tab1} presents the coding scheme along with the definition for each code, designed for analyzing university policies and statements.

\begin{table*}[ht]
\caption{Coding scheme for analyzing guidelines and resources provided by the universities.}\label{tab2}
\begin{tabularx}{\textwidth}{@{}TVY@{}}
\toprule
Parent Codes                            & Child Codes                & Definition\\ \midrule
\multirow{4}{*}{Target Audience}       & For Students & Resources provided specifically for students.\\ \cmidrule(l){2-3}
                                        & For Faculty & Resources provided specifically for faculty.\\ \cmidrule(l){2-3}
                                        & For General Audience & Resources provided specifically for the wider university community and public.\\ \midrule
\multirow{14}{*}{Types of Resources}     & Syllabus Templates and/or Examples & Suggested syllabus templates and/or examples shared on the resource and guideline pages.\\ \cmidrule(l){2-3}
                                        & Practical Training Workshop  & Training workshops that train instructors and/or students to learn and try various functions of GenAI.\\ \cmidrule(l){2-3}
                                        & Dialogues and Discussions & Open dialogues and discussions for instructors and/or students to share their opinions.\\ \cmidrule(l){2-3}
                                        & Shared Articles and/or Blogs & Referenced articles and/or blogs that help instructors and/or students to further explore relevant topics.\\ \cmidrule(l){2-3}
                                        & One-on-one Consultations & Individual email, Zoom, and/or in-person consultations with school administration offices or representatives.\\ \midrule
\multirow{17}{*}{Content Analysis}       & General Technical Introduction & An overview focusing on the functions and technical aspects of GenAI. \\ \cmidrule(l){2-3}
                                        & Ethical Considerations & An introduction of ethical concerns on the use of GenAI. \\ \cmidrule(l){2-3}
                                        & Pedagogical Applications & Exploration of how GenAI can be incorporated into teaching and learning. \\ \cmidrule(l){2-3}
                                        & Preventive Strategies & Strategies to prevent students from using GenAI inappropriately. \\ \cmidrule(l){2-3}
                                        & Data Privacy & Guidelines for protecting instructors’ and students’ privacy when using GenAI for teaching and learning. \\ \cmidrule(l){2-3}
                                        & Limitations & Concerns on limitations, including biased, inaccurate, unreliable, or falsely cited information generated by AI. \\ \cmidrule(l){2-3}
                                        & Detective Tools & Introduction of available detective tools for detecting the use of GenAI. \\ \bottomrule
\end{tabularx}
\end{table*}

Table \ref{tab2} shows the codes and definitions designed for analyzing resources and guidelines provided by the universities regarding the use of GenAI.

After collecting data, the primary researcher thoroughly examined part of the data and then induced the initial child codes and broader parent codes. The codes were presented along with definitions in a table and introduced to other researchers in this study. Then other researchers reviewed and finalized the two coding schemes. Next, the coding schemes have been applied to the entire data for comprehensive analysis. The researchers had regular meetings to verify the coding results and discuss discrepancies.

\subsection{Scale and Point Systems}
\label{sec4-3}
This study further delves into how school ranking tiers and academic specializations shape the trends in perceptions and resource provision for GenAI in higher education. For this purpose, we developed a scale system in order to quantify different universities’ perceptions, from proactive embrace to cautious hesitance (see Table \ref{score}). The perception points are assigned according to each university's policy stances, with '0' representing 'Undecided/Unclear' policies to reflect a neutral and open position towards the use of GenAI. Various negative and positive scores in the range of [-5, 5] illustrate the spectrum from cautious hesitance to strong endorsement respectively.

\begin{table}[h]
\caption{University perception scale on AI usage.}\label{score}
\begin{tabular}{@{}lc@{}}
\toprule
University policies and decisions & Scale points\\
\midrule
Ban use    & -5  \\
Undecided/Unclear    & 0  \\
Allow use with conditions    & 5  \\
Instructor decides (prohibition by default)    & 1  \\
Instructor decides (neutral)    & 2.5  \\
Instructor decides (permissibility by default)    & 4  \\
\bottomrule
\end{tabular}
\end{table}

Another scoring system extends to quantify the comprehensiveness of resources provided by these institutions. This is achieved by evaluating the breadth and depth of the resources, considering the target audience, the variety of resource types, and the range of content categories provided in Table \ref{tab2}. Specifically, universities accumulate scores based on their available resources across the three dimensions. For instance, for the target audience, a university earns one point for each distinct group for whom resources are provided, including students, faculty, or the general audience, with a maximum of three points available in this dimension. For the types of resources, a university gains one point for each type they provide, such as syllabus templates and workshops, allowing for up to five points in this dimension. Similarly, regarding the content categories, universities are awarded one point for each covered topic or theme, such as technical introductions and ethical considerations, for up to seven points for this dimension. Both the quantitative scale and point systems are used specifically in Section \ref{sec5-3} for exploring how key variables (ranking and academic specialization) affect the trends in perceptions and resource provision. 

\begin{figure*}[t]
\centering
\includegraphics[width=0.9\textwidth]{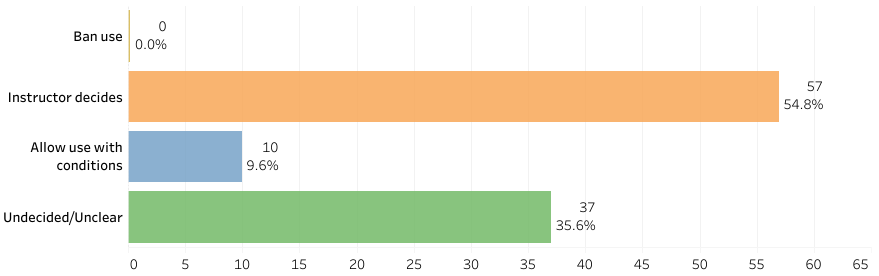}
\caption{Policies and stances adopted by different universities regarding GenAI.}\label{horiz_1}
\end{figure*}

\section{Results}\label{sec5}

\subsection{Policies, Management Approaches, and Perceptions from U.S. Universities } 

To answer RQ1, this section introduces the current policies, and management approaches adopted by the top-ranked U.S. universities, as well as the perceptions and implications derived from these policy statements. Fig. \ref{horiz_1} summarizes policies and stances adopted by the top 100 universities (n = 104) in the U.S. regarding the legality and application of ChatGPT and other GenAI tools in higher educational contexts. None (n = 0, 0\%) of the top 100 universities have completely banned these tools, reflecting a general acceptance or openness towards GenAI. The majority (n = 57, 54.8\%) give this decision-making agency to individual instructors, indicating a contextualized and faculty-centric approach. Meanwhile, a modest 9.6\% (n = 10) had implemented conditional use policies with proper citations, and 35.6\% (n = 37) remained either undecided or had not clearly announced their policies or stance. These responses demonstrate a diverse but flexible approach to integrating AI in higher education contexts in general.

However, it should be highlighted that no decision or no clear policy does not imply indifference toward GenAI on the part of these universities. Instead, many of them often present an open and objective introduction of ChatGPT and/or other AI writing tools, which represent their neutral perceptions. For example, the University of Illinois at Urbana-Champaign outlines both the benefits and challenges of using GenAI and advises that careful thoughts and considerations should be kept in mind to incorporate GenAI into coursework. While the universities are not refusing AI tools completely, they often stand in a neutral position and share all resources in a balanced way. 

\begin{figure*}[t]
\centering
\includegraphics[width=0.8\textwidth]{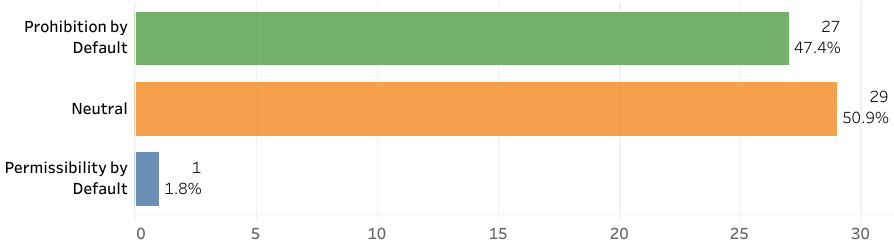}
\caption{Default policies when universities encourage instructors to decide.}\label{horiz_2}
\end{figure*}

When selecting the "instructor decides" policy towards GenAI, a more cautious trend emerges, as illustrated in Fig \ref{horiz_2}. Among the 57 “instructor decides” universities, 27 (47.4\%) adopt a stance of Prohibition by Default, only allowing the use of such tools when an instructor explicitly permits it. If the instructor has not presented any policy statements on the use of GenAI, using ChatGPT in homework and essays is generally not allowed and may be under the circumstance of plagiarism. If the instructor allows it, students must cite appropriately and take responsibility for their responses. This option reveals the universities’ more cautious perceptions and evident concerns about the impact of GenAI on academic integrity. 

29 (50.9\%) universities adopt a more open and neutral stance, granting instructors the autonomy to decide and addressing transparency in policy-making. This approach signals a more balanced and pragmatic perspective towards GenAI tools and also reflects the universities’ practical considerations and respect for the diverse needs and contexts of different disciplines. University of California, Irvine (UCI)’s statement serves as an example, showing the essence and rationale behind this approach.

\textit{``Individual faculty will need to make decisions based on the context of their course, course objectives, students' academic progression, and disciplinary-specific goals of their students' learning experiences''} (UCI Generative AI for Teaching and Learning\footnote[2]{https://dtei.uci.edu/chatgpt}).

\begin{figure}[t]
\centering
\includegraphics[width=\columnwidth]{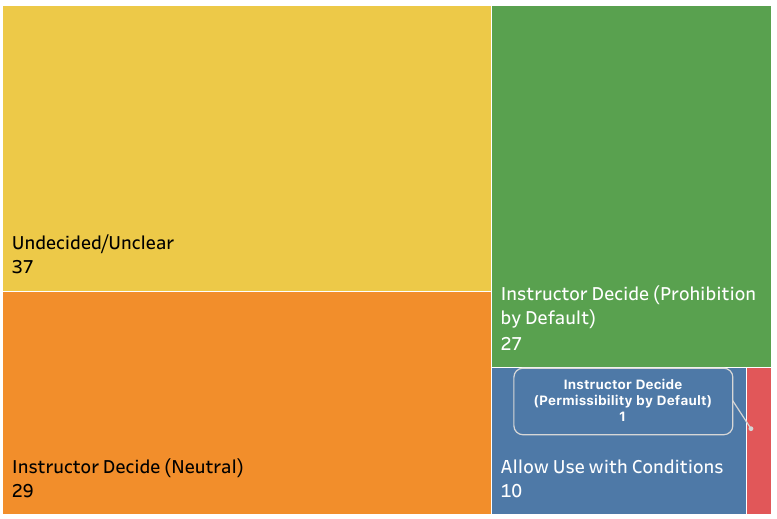}
\caption{Policies from the top 100 US universities regarding GenAI.}\label{rect}
\end{figure}

In summary, the different approaches of the top 100 universities on the use of ChatGPT (see Fig. \ref{rect}) illustrate that US top universities tend to show a generally open but cautious stance with a strong tendency towards encouraging instructors to manage the use of ChatGPT according to their own teaching contexts. The diversity can also reflect the uncertainty and complexity of adapting AI in higher education.

\begin{figure}[t]
\centering
\includegraphics[width=\columnwidth]{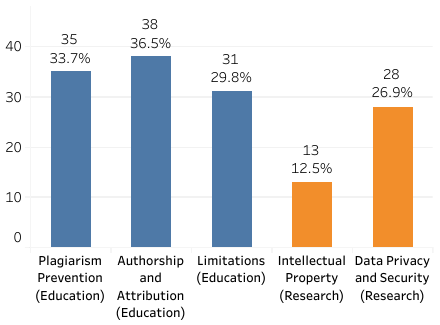}
\caption{Purposes and focuses of the policies.}\label{purpose}
\end{figure}

We are also interested in exploring the main topics covered in the existing policies concerning GenAI technologies. Fig. \ref{purpose} introduces the purposes and focuses of the existing policies on the use of ChatGPT in higher education. The data reveals a focus on addressing educational challenges and concerns, with higher attention to issues such as plagiarism (n = 35, 33.7\%), inadequate proper attribution and citations (n = 38, 36.5\%), and the limitations of AI tools (n = 31, 29.8\%). On the other hand, topics related to professional research writing, such as intellectual property (n = 13, 12.5\%) and data privacy (n = 28, 26.9\%), have received comparatively less attention in these policies. This trend reveals that policy development in higher education institutions across the U.S. may often pay more attention to educational areas. Professional research writing and publication in academia may need more guidelines from institutions.

\begin{figure}[ht]
\centering
\includegraphics[width=0.55\columnwidth]{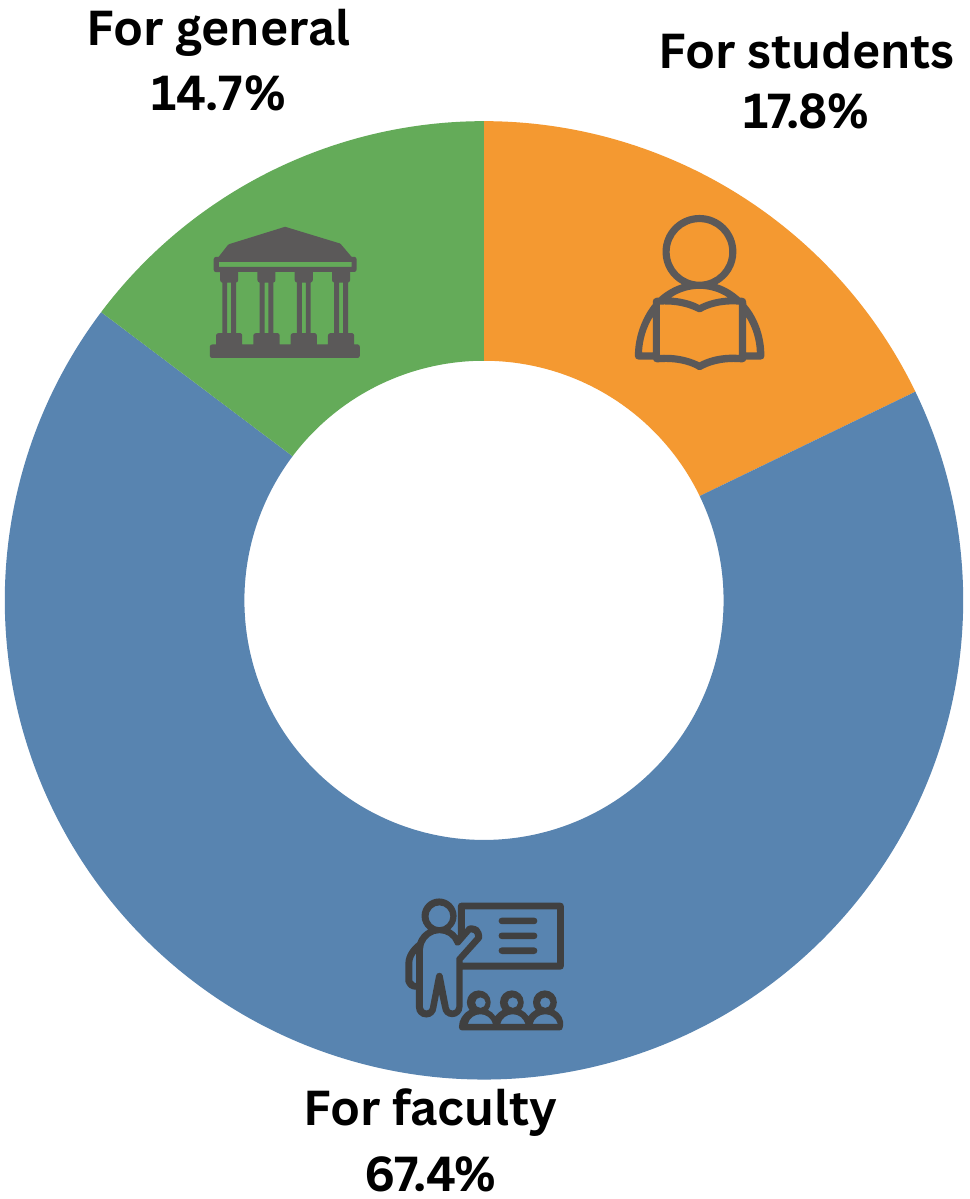}
\caption{Policies from the top 100 US universities regarding GenAI.}\label{audience}
\end{figure}

\subsection{Guidelines and Resources for the Applications of GenAI} 
RQ2 asks about resources and guidance that U.S. universities provide regarding the use of GenAI tools in higher education contexts. This section reports the target audience, types of resources, and content analysis with a focus on limitations, pedagogical applications, and prevention strategies.

\subsubsection{Target Audience} 
In regards to the guidelines and resources about ChatGPT and other GenAI, we first analyzed their aimed target audience (see Fig. \ref{audience}). The results show that 70 (67.4\%) universities of the top 100 universities across the U.S. have resources and guidelines explicitly designed for faculty and instructors. A smaller portion of 19 (17.8\%) universities offer resources aimed at students, and 15 (14.7\%) provide guidance for the broader audience, including faculty, students, and staff, without specifying a particular audience. The findings show a broader emphasis on resources crafted for the population of faculty and instructors to incorporate GenAI tools into their teaching practice. Relatively fewer resources address and guide students on the appropriate application of GenAI tools in their learning. This discrepancy highlights the demands to further develop in-depth guidelines for students and establish a more inclusive GenAI resource and support system.

\subsubsection{Types of Resources} 
\begin{figure}[t]
\centering
\includegraphics[width=\columnwidth]{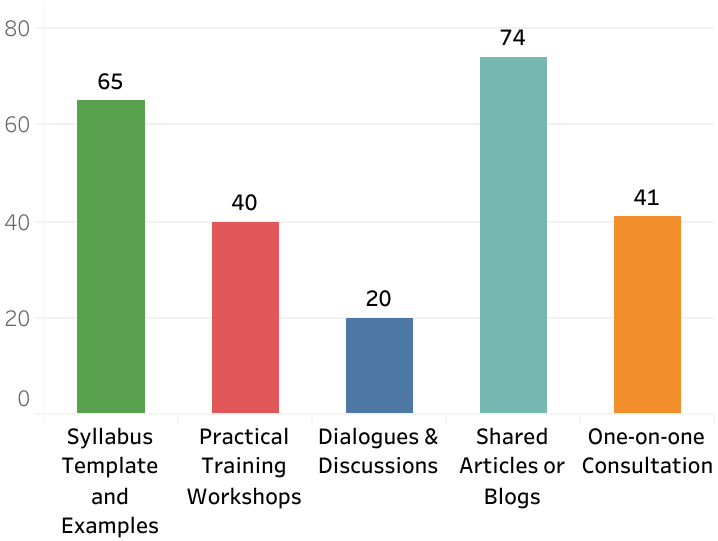}
\caption{Types of resources regarding ChatGPT provided by the top 100 universities.}\label{type}
\end{figure}

Fig. \ref{type} illustrates the diversity of resources from the top 100 universities (n = 104), highlighting Shared Articles or Blogs being the most prevalent type (n= 74, 71.1\%). The resource centers often feature additional reading at the bottom of the pages and/or embedded links within the content. The resources include research papers, news articles, other university websites, and blogs and cover a wide range of topics, such as opportunities and challenges of GenAI for education, educators’ and students’ reactions to GenAI innovations, and teaching strategies with the use of ChatGPT. These resources are important for instructors to gain foundational knowledge about emerging technologies. 

\begin{figure}[t]
\centering
\includegraphics[width=\columnwidth]{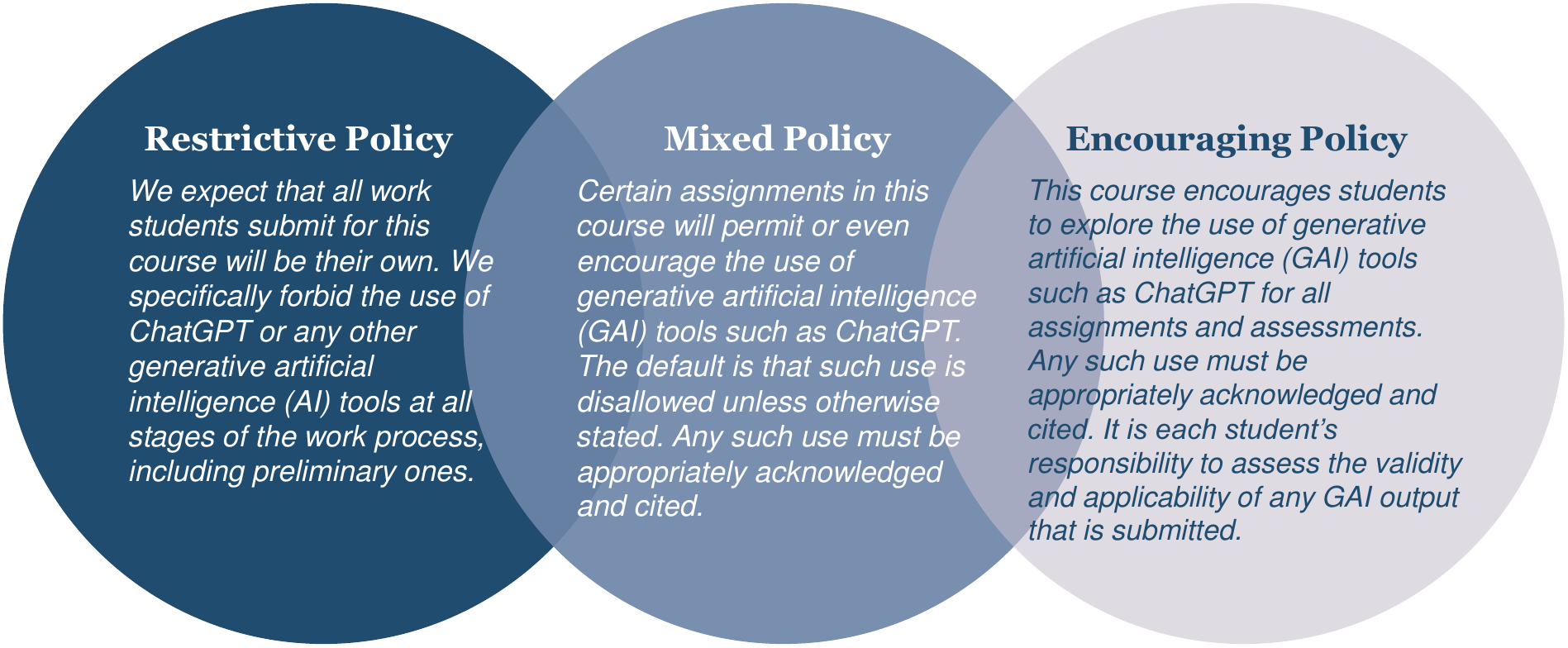}
\caption{Syllabus samples from Harvard University.}\label{Harvard}
\end{figure}

It appears from our data that 65 (62.5\%) universities have offered syllabus templates and examples as references helping instructors and teacher trainers make policy decisions according to their own teaching contexts. These templates typically showcase three distinct policy perspectives—restrictive, mixed, and encouraging, to accommodate diverse disciplines’ contexts. Harvard University's syllabus samples\footnote[3]{https://oue.fas.harvard.edu/ai-guidance} exemplify this approach (see Fig. \ref{Harvard}). Universities usually encourage all instructors to explicitly include a clear policy in course syllabi regarding the use and misuse of ChatGPT and other GenAI tools. Open and explicit communications are crucial to help students understand the boundaries and expectations when they interact with GenAI tools in learning.

Additionally, 41 (39.4\%) universities include one-on-one consultations as a type of resource for instructors and/or students. They are conducted with the institution's GenAI specialists and/or representatives from the Teaching and Learning Center to address attendees’ specific concerns and navigate the personal applications of ChatGPT and other GenAI tools in their own teaching and learning contexts. 

Furthermore, it is evident in Fig. \ref{type} that a smaller proportion of universities offer workshops (n = 40, 38.5\%) and discussions (n = 20, 19.2\%) regarding the use of ChatGPT and other GenAI tools. This might be because many of these workshops and discussions are a part of internal resources and are not publicly available. As ChatGPT is a new emerging technology, workshops and discussions are crucial for familiarizing educators with its positive applications and educational implications \citep{de2023chatgpt}. The results could reflect that there is a possible demand for increasing the frequency and accessibility of these events.

\subsubsection{Content Analysis of Resources and Guidelines} 
\begin{figure}[t]
\centering
\includegraphics[width=\columnwidth]{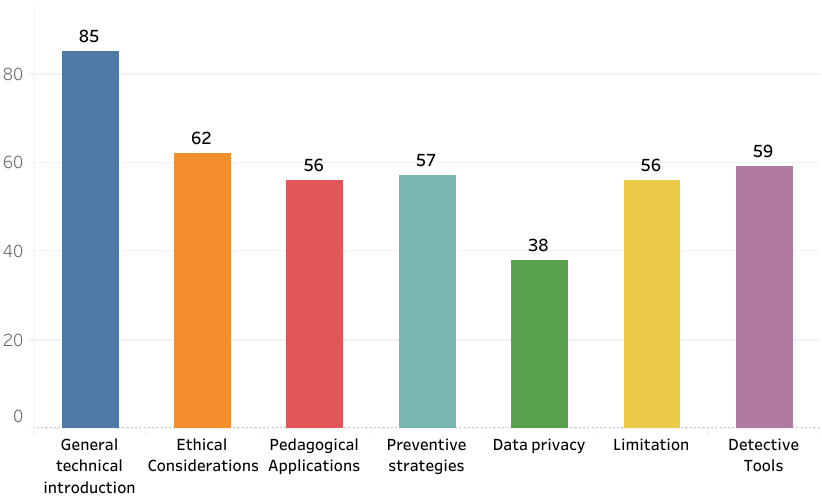}
\caption{Content analysis of resources and guidelines from the top 100 universities.}\label{content_anly}
\end{figure}

Fig. \ref{content_anly} summarizes the focuses and purposes of the existing resources and guidelines regarding the use of ChatGPT and other GenAI tools in higher education. The majority of the universities start by introducing some general technical information (n = 85, 81.7\%), such as a beginning section named What is ChatGPT? or GenAI in Education. This approach indicates the institutions' intention to familiarize faculty, students, and staff with GenAI tools and enhance their understanding of these technologies. 62 (59.6\%) universities discuss the ethical implications of implementing ChatGPT in higher education, including their apprehensions about the misuse of GenAI to foster plagiarism, violate academic integrity, and negatively affect student evaluation. This discussion highlights the necessity for instructors to monitor and guide the use of ChatGPT \citep{huallpa2023exploring}, as well as the need to adapt teaching methodologies to prevent possible cheating and plagiarism.

56 (53.8\%) universities explicitly list the inherent limitations of the current GenAI model and 38 (36.5\%) universities address the data privacy issue, which serves as a reminder for individuals to keep in mind that these constraints may influence teaching approaches, learning experiences, and the conduction of research. Being aware of these limitations and issues ensures that GenAI tools would be used in a more responsible and ethical manner in education \citep{kasneci2023chatgpt}. The limitations primarily include: 

\begin{itemize}
    \item Inaccuracy or misleading information;
    \item Biased opinions based on the training data;
    \item Fake information and/or hallucinations: especially when generating citations and references;
    \item Limited knowledge of recent information and specific academic fields;
    \item Absence of citations and references.
\end{itemize}

The inevitable intertwining with GenAI tools needs teachers and educators to evolve the teaching methods across all subjects. Therefore, numerous top 100 universities have offered resources on pedagogical applications (n = 56, 53.8\%) and prevention strategies (n = 57, 54.8\%). They cover the effective incorporation of ChatGPT into classrooms to enhance student learning experiences, alongside strategies to avoid its misuse by students. Table \ref{app} provides a summary of major pedagogical applications and prevention strategies sourced from these leading universities. Both approaches highlight the significance of cultivating students' critical thinking skills and problem-solving abilities, which are essential competencies in today’s rapidly advancing, technology-infused world.

\begin{table*}[t]
\caption{Pedagogical applications and prevention strategies of AI in teaching and learning.}\label{app}
\begin{tabularx}{\textwidth}{@{}XX@{}}
\toprule
\makecell{\textbf{Pedagogical Applications} \\ Effectively incorporate AI \\ in teaching and learning} & \makecell{\textbf{Prevention Strategies} \\ Prevent inappropriate use of AI \\ in teaching and learning}\\
\midrule
\begin{itemize}
  \item Ask students to analyze and evaluate AI-generated texts. 
  \item Ask students to compare and evaluate the different versions of texts generated by different AI tools.
  \item Ask students to compare/contrast AI-generated texts with human writing.
  \item Ask students to revise and edit AI-generated information.
  \item Ask students to debate or argue with AI and reflect on their learning.
  \item Use AI as a resource for students to receive feedback on their drafts
  \item Use AI tools to brainstorm initial teaching ideas and activities. 
  \item Use AI tools to generate additional examples of certain concepts. 
  \item Use AI tools to summarize long or difficult text. 
  \item Use AI tools to generate writing prompts, grading rubrics or quiz questions based on the course materials. 
  \end{itemize}
  &
  \begin{itemize}
  \item Ask students to explain their thought processes as they solve problems.
  \item Ask students to reflect on their personal learning experiences and opinions. 
  \item Ask students to connect with their personal knowledge and life experiences.
  \item Ask students to include and provide proper academic citations.
  \item Ask students to reference class materials, notes, or sources that are unavailable online.
  \item Ask students to complete assignments in class.
  \item Ask students to present their answers in multimodal ways, such as hand drawing, or audio threads. 
  \item Include visual prompts in assignments.
  \item Design assignments related to current events or discussions in the specific academic field.
  \item Divide the larger project into multiple smaller tasks. 

\end{itemize}\\
\bottomrule
\end{tabularx}
\end{table*}

Another trend that emerges from the data is the discussion of using GenAI detection tools or GenAI detectors to identify AI-generated text in students’ work. 59 (56.7\%) universities discuss the available common detective tools, such as Turnitin and GPTZero. However, it is worth noting that none (n = 0, 0\%) of the universities in this study view the use of detective tools as a completely reliable method to identify AI-generated information, and none (n = 0, 0\%) of them support instructors to use of the tools to evaluate students’ academic integrity and determine plagiarism. While GenAI detectors are designed to identify AI-generated language patterns, the research conducted by \citet{sadasivan2023can} shows that they are not reliable in many real-world scenarios. This is particularly evident when the detective tools are faced with paraphrasing attacks which refer to applying a light paraphraser to generated texts. Even a minor rephrasing can significantly affect the performance and accuracy of the entire detection system \citep{sadasivan2023can}. Additionally, universities in this study raise further concerns regarding the use of GenAI detectors, including the potential violation of students' intellectual property rights and the risk to data privacy once their work is submitted to the detective tools. Some universities also believe that relying on such tools might undermine the relationship of trust between students and teachers as well.

\subsection{Trends in Perceptions and Resource Provision of GenAI}
\label{sec5-3}
RQ3 examines how universities’ perceptions and resources on the use of GenAI are affected by two possible factors: 1) institution ranking tiers; and 2) institution academic specializations. We will discuss the pedagogical implications that can be drawn from these findings. 

\subsubsection{Perception and Available Resources Across Different Tiers} 
This section employs the scale and point system described in method Section \ref{sec4-3} to examine the relationship between three dimensions, including perceptual stances, resource diversity, and university rankings. We divide the top 100 universities (n = 104) into three tiers to examine the disparities in perceptions and resources across different tiers of higher education institutions. Tier 1 includes the top 1-33 universities (n = 34); Tier 2 comprises the subsequent top universities ranked 34-66 (n = 32); and Tier 3 includes the remaining reputational universities ranked 67-100 (n = 38). 

\begin{figure}[t]
\centering
\includegraphics[width=\columnwidth]{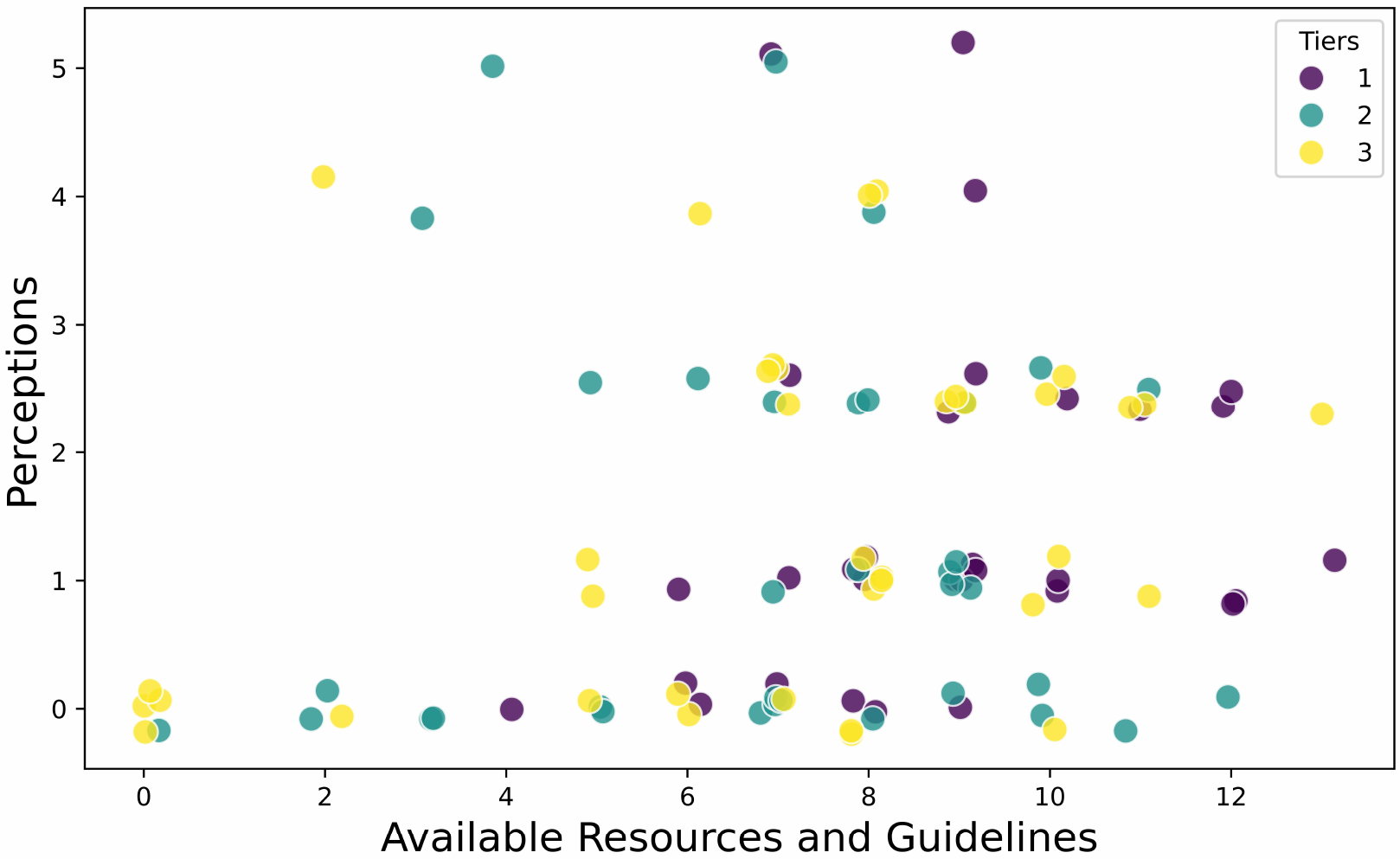}
\caption{Perception scores vs. available resources across different tiers.}\label{tier}
\end{figure}

In terms of the results, Fig. \ref{tier} illustrates the correlation between the perceptions and resources regarding GenAI across the three ranking tiers. We find that a majority of the top 100 universities have developed a diversity of resources and guidelines for integrating GenAI, demonstrating a proactive approach towards applying this technology to their education. However, the figure shows that there is no significant correlation between the universities' rankings and the depth of perceptions and resources related to GenAI. The trend also indicates that a large number of universities present unclear and cautious perceptions toward GenAI. These perceptions might be due to uncertainties and controversial features of GenAI in higher education.

\begin{figure}[t]
\centering
\includegraphics[width=\columnwidth]{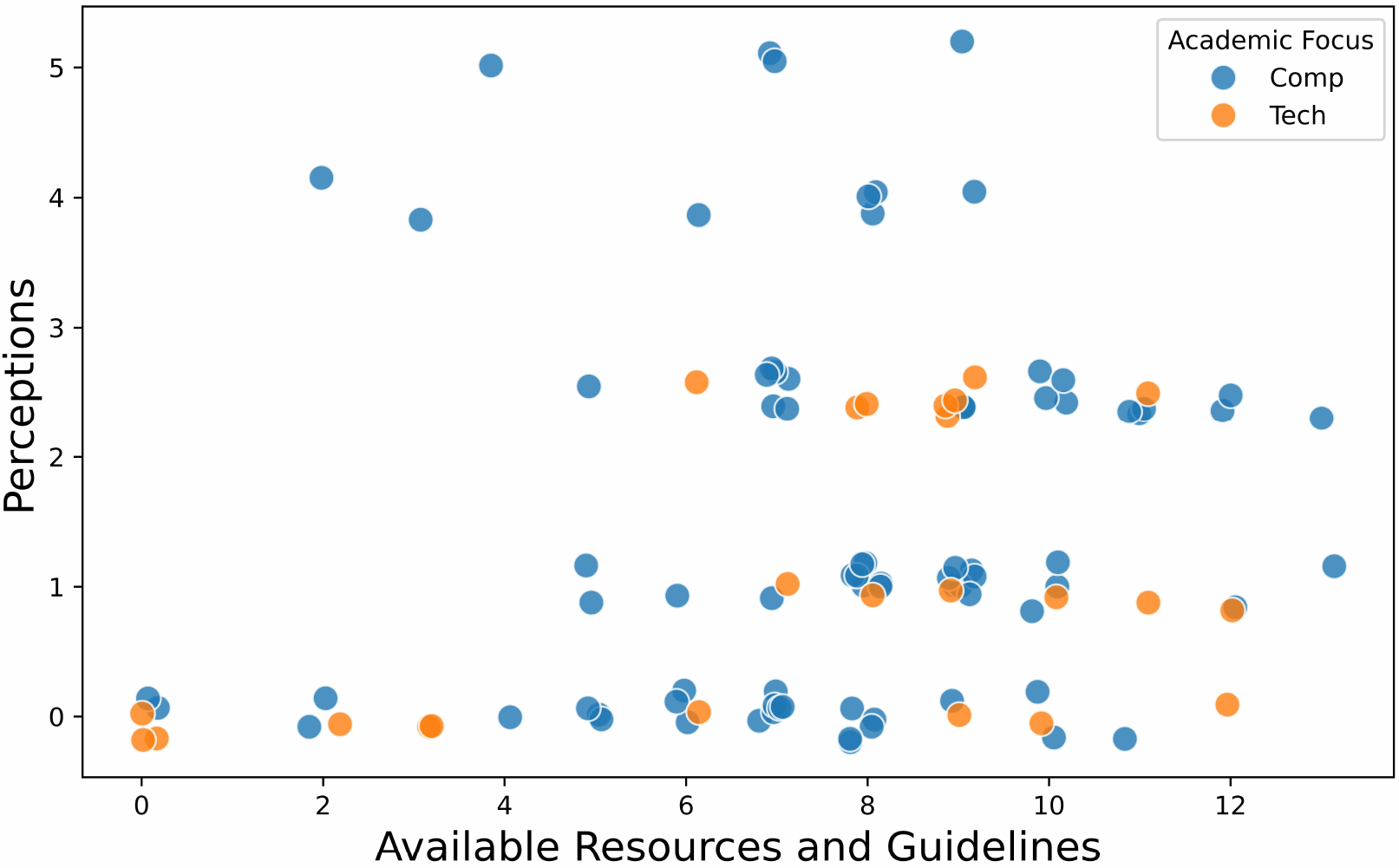}
\caption{Perception and available resources across different types of academic focuses.}\label{focus}
\end{figure}

\subsubsection{Perception and Available Resources Across Different Academic Focuses} 
We have also examined how different academic focuses of the universities affect their perceptions and resource provision regarding GenAI tools. We divide the top 100 universities (n = 104) into two groups. Institutions recognized for their programs in technology, engineering, and science, such as the California Institute of Technology, are categorized as “technology-oriented universities” (n = 24). Conversely, universities known for their broader and comprehensive academic subjects, covering areas such as arts, social sciences, and humanities alongside other fields are classified as “comprehensive universities” (n = 80), such as Harvard University. 

An evident trend emerges when comparing the perceptions and resources across the two groups (see Fig. \ref{focus}). The technology-oriented universities show a higher level of caution and careful consideration in their policies along with more comprehensive and diverse guidelines and resources, which reflect an active engagement with GenAI’s intricacies and potential complex implications in their academic contexts. For example, programming homework can be prevalent in some STEM courses, so there might be more concerns about academic dishonesty due to GenAI's ability to provide definitive solutions \citep{michel2023challenges}. On the other hand, there are a number of comprehensive universities that tend to adopt a more welcome, supportive, and positive stance regarding the use of GenAI. This variation may indicate that the academic specialization of a university might affect its approach to managing and integrating GenAI in its academic education settings. Considering the nature of different academic disciplines and contexts, the results highlight the importance of developing discipline-specific policies and guidelines that can address challenges and accommodate the needs of various academic domains. It can also reflect that the integration of GenAI in educational practice needs to align with the objectives of each discipline to effectively enhance student learning experiences.

\section{Discussion}\label{sec6}
The integration of ChatGPT and other GenAI tools into educational contexts has been mixed with enthusiasm and concerns. In order to incorporate GenAI tools in higher education in a more ethical and responsible way, this study analyzes the currently available policies, statements, guidelines and resources regarding GenAI, especially ChatGPT,  of the top 100 universities in the U.S.. The results show that most of the universities tended to approach the situation with careful consideration and a rich diversity of teaching support, which reflects the increased awareness and efforts from higher education institutions. Some were in a period of observation, evaluating more impacts of GenAI on the educational contexts or waiting for the approval of updated university policies \citep{sullivan2023chatgpt}.

\subsection{Implications for Educators in Teaching Practices} 
For educators in teaching practice, the study offers a comprehensive overview of the primary resources regarding using GenAI in higher education. Despite the potential risks posed by GenAI, we find that numerous universities have introduced the potential benefits of GenAI and proposed pedagogical applications that enable instructors to leverage ChatGPT effectively in their teaching preparation and practice. With appropriate guidance, GenAI can serve as a helpful and powerful tool for teachers in activity development, implementation, and assessment \citep{hodges2024innovation, mishra2023tpack, oravec2023artificial}. Given the practical impossibility of prohibiting ChatGPT use among university students \citep{sullivan2023chatgpt}, teachers and educators have come to accept, adapt, and embrace its presence \citep{moorhouse2023generative} and actively engage with university resource centers, seeking new techniques and approaches to incorporate GenAI to enhance student learning. 

As a majority of universities give instructors agency to regulate and incorporate the use of GenAI in their own classes, it is important for instructors to decide and apply to align with their subjects’ specific contexts and learning objectives. The finding offers major pedagogical strategies that can be insightful for teaching with ChatGPT, such as using GenAI tools to generate writing prompts, grading rubrics, or quiz questions based on the course materials, asking students to compare/contrast AI-generated texts with human writing, and asking students to revise and edit AI-generated information. To incorporate GenAI into classrooms, we suggest instructors explore the pros and cons of GenAI and consider how it can enhance or detract from their subject-specific teaching environments. It is crucial to reflect on the specific student learning outcomes and how GenAI can be used to achieve these goals. For example, in a genre-based writing class, ChatGPT might be employed as a tool for providing writing samples for different contexts, purposes, and audiences, thereby demonstrating different writing conventions and tones. Conversely, in a course aimed at developing research skills, instructors might advise against dependence on ChatGPT for providing article sources, emphasizing instead the importance of students’ critical evaluation of sources and original thought. 

To address the concerns about plagiarism and academic dishonesty in teaching and learning, teachers should consider updating their curriculum and evolving activities to avoid students’ misuse of GenAI. First, we suggest instructors establish clear policies and guidelines to explicitly specify to students what they can and cannot do with GenAI in the course syllabi. Given that numerous universities have already offered templates or examples for syllabus language, developing the policies and guidelines should be straightforward. Second, having in-class discussion activities about the ethical use of GenAI and its impact on academic integrity can also help students understand the difference between collaboration with GenAI and plagiarism. Third, instructors can consider evolving their curriculum and activities by applying some of the prevention strategies discussed in the findings, such as asking students to connect with their personal knowledge and experiences, providing proper academic citations, and referencing class materials that are unavailable online. It would also be helpful for instructors to divide their assignments into smaller steps (brainstorming, first draft, second draft, peer review, reflection) to allow students to engage in learning throughout the course instead of relying on one big submission at the end of their course. The discussion and activities can not only prepare students to navigate the complexities of technology but also promote a deeper understanding of the responsibilities that come with using such tools.

Last but not least, considering the current GenAI detection tools are not reliable and supported to determine plagiarism by most universities \citep{elkhatat2023evaluating, weber2023testing}, teachers should adopt multifaceted evaluation strategies to assess students’ work. One practical approach can be to assess students' written work based on their previous and in-class performance. AI-generated content cannot fully replicate the unique writing style of individual students, which often includes word selection, phrase usage, language patterns, and students' personal insights developed over their learning. Comparing multiple works for consistency may be a way to identify AI-generated content. Some other prevention strategies, such as asking students to explain their answers in multimodal ways, such as presentations and audio threads, discussed in the results section can also be applied to avoid the inappropriate use of GenAI. It is important to note that instructors should treat each student’s work as a learning opportunity, thus, providing feedback at multiple levels and asking students to reflect on how they make changes to their assignments can be beneficial in both areas (preventing the excessive use of GenAI and students’ gaining knowledge). In addition, accusing students of using GenAI needs to be addressed with caution and care, in order to maintain a trusting relationship between students and their instructors. More guidance on how to establish these harder conversations between faculty and students needs to be discussed and developed further. 
 
\subsection{Recommendations for Educators to Make Policies and Guidelines} 
For teacher trainers and educators who make policies and provide guidelines, this study suggests that it is crucial to establish clear policies and guidelines with the consideration of discipline-specific contexts. While many universities allow instructors to decide their own course policies regarding the use of GenAI, it can be challenging for instructors as they have to not only understand the needs of their students but also align that with the ongoing changes of these tools 
\citep{chiu2023impact, zastudil2023generative}. To help with establishing policies and guidelines, policymakers can engage with educators across various departments to understand how they currently teach and prepare to teach, what assignments and activities their students are working on, and areas they need to be aware of. Developing policies and resources should be a shared effort that is designed with consideration of a range of academic domains. Depending on some programs’ special contexts and needs, teacher trainers and educators should be encouraged to discuss, evaluate, shape, and refine the policies to ensure the rules align with and support the specific needs of their teaching contexts.

The results also recommend taking precautions when managing sensitive or proprietary information, whether faculty their own or that of their students. Teaching resources and teacher trainers can prepare more training workshops to explicitly discuss the inherent privacy risks of GenAI and how to protect students’ personal information if ChatGPT is integrated into teaching, especially in some activities such as grading and providing feedback. Guidelines and resources should clearly outline what types of information that are safe to share with GenAI and what types are not. On the other hand, academic integrity should extend beyond teaching to include the principles, behaviors, and ethics observed in research as well \citep{macfarlane2014academic}. More explicit policies and guidelines are necessary to raise researchers' awareness of which information is considered sensitive and personal and to address the appropriate boundaries for using GenAI in research in higher education.
 
\section{Conclusion}\label{sec7}
This study delves into the academic policies, resources, and guidelines of the top 100 U.S. universities (n = 104) regarding ChatGPT and other GenAI tools in higher education, and thereby informs educators in teaching practices as well as future policy-making. Data was collected from publicly available official university sources, such as the Office of Provost and the Center of Teaching and Learning. The results reveal a prevalent balanced and open yet cautious and thoughtful attitude toward integrating GenAI technology given concerns mainly on ethical issues, inherent limitations, and data privacy. A number of universities encourage instructors to develop their own policies and guidelines for the use of GenAI (n = 57, 54.8\%), respecting the specific contexts and needs of their disciplines. For resources and guidelines, most popular teaching support includes syllabus samples and templates (n = 65, 62.5\%), workshops (n = 40, 38.5\%), articles (n = 74, 71.1\%), and individual consultations with topics on technical introduction (n = 41, 39.4\%), pedagogical applications (n = 56, 53.8\%), prevention strategies (n = 57, 54.8\%), limitations (n = 56, 53.8\%), and detection tools (n = 59, 56.7\%), to help instructors adapt their teaching practices in the age of GenAI. These efforts highlight a variety of opportunities and challenges presented by GenAI along with raising the necessity to further enrich and refine the current guidelines as well as curriculum in higher education.

In the era of GenAI, what we should not do is to stay stagnant. Actively engaging with these technology advancements is significant for better leveraging their potential and effectively mitigating their risks. The findings of this study may have important implications for educators in the contexts of both teaching practices and policy design. For educators in teaching practices, the pedagogical implications include accepting, adapting, and embracing the presence of GenAI, aligning its use with specific learning objectives, updating the curriculum to guide and prevent students from misuse, as well as applying multifaceted evaluation strategies instead of relying on GenAI detectors. For educators who need to make policies for their own classes and/or departments, we recommend designing policies according to their discipline-specific contexts and take precautions when managing sensitive information.

\section*{Declarations}



\textbf{Declaration of competing interest:} The authors declare no financial or personal relationships that could inappropriately influence their work.

\textbf{Data availability:} The authors declare that the data supporting the findings of this study are available within the paper.

\textbf{Funding:} This research did not receive any specific grant from funding agencies in the public, commercial, or not-for-profit sectors.

\textbf{Acknowledgements:} Not applicable.

\bibliographystyle{cas-model2-names}

\bibliography{cas-refs}


\end{document}